\documentclass{article} 
\usepackage{nips14submit_e,times}
\usepackage{hyperref}
\usepackage{url}
\usepackage{wrapfig}

\usepackage{hyperref}
\usepackage{amssymb}
\usepackage{amsmath}
\usepackage{graphicx} 
\usepackage{subfigure}

\usepackage{times}

\usepackage{algorithm}
\usepackage{algorithmic}
\usepackage{amssymb}
\usepackage{amsmath}
\usepackage{multicol,array}
\usepackage{bm}

\DeclareMathOperator*{\argmin}{arg\,min}

\DeclareMathOperator*{\minimize}{minimize}

\newcommand{\AAA}{{\mathcal{A}}}
\newcommand{\DD}{{\mathcal{D}}}
\newcommand{\EE}{{\mathcal{E}}}

\newcommand{\GG}{{\mathcal{G}}}
\newcommand{\HH}{{\mathcal{H}}}
\newcommand{\LL}{{\mathcal{L}}}
\newcommand{\TT}{{\mathcal{T}}}
\newcommand{\QQ}{{\mathcal{Q}}}
\newcommand{\RR}{{\mathcal{R}}}
\newcommand{\XX}{{\mathcal{X}}}
\newcommand{\YY}{{\mathcal{Y}}}

\usepackage{algorithm}
\usepackage{algorithmic}
\usepackage{float}

\begin{document}

\title{Representation as a Service}
%

%
\author{
Ouais Alsharif, Philip Bachman, Joelle Pineau\\
School of Computer Science\\
McGill University \\
Montreal, Canada\\
}

\nipsfinalcopy
\maketitle              

\begin{abstract}

Consider a Machine Learning Service Provider (MLSP) designed to rapidly create highly accurate learners for a never-ending stream of new tasks. The challenge is to produce task-specific learners that can be trained from few labeled samples, even if tasks are not uniquely identified, and the number of tasks and input dimensionality are large.  In this paper, we argue that the MLSP should exploit knowledge from previous tasks to build a good representation of the environment it is in, and more precisely, that useful representations for such a service are ones that minimize generalization error for a new hypothesis trained on a new task. We formalize this intuition with a novel method that minimizes an empirical proxy of the intra-task small-sample generalization error. We present several empirical results showing state-of-the art performance on single-task transfer, multitask learning, and the full lifelong learning problem.

\end{abstract}
\pagenumbering{arabic}

\section{Introduction}
\label{chap:Introduction}

Consider a Machine Learning Service Provider (MLSP) designed to rapidly create highly accurate learners for a never-ending stream of new tasks. Different clients, for example a social networking site or video surveillance company, could ask the MLSP to design a stream of different face recognition agents, each achieving recognition of a different set of target individuals.  In such a setting, it is necessary to quickly produce a task-specific learner that can be trained from very few labeled samples (e.g. examples of the target face).   Learning from few labeled samples has been known to arise in many tasks for which data is expensive to label (e.g., medical images) or slow to collect (e.g., human computer interaction). In general, there are three paths towards acceptable performance for learning from few samples: 
\vspace{-\topsep}
\begin{enumerate}
\itemsep 0em
\item Using domain knowledge.
\item Using unlabeled samples.
\item Using labeled samples from a different, but related task.
\end{enumerate}
\vspace{-\topsep}
Bayesian methods \cite{fei2007learning} have typically focused on the first option, using knowledge of structure in the target task to bias search towards better hypotheses. Other methods, like manifold regularization \cite{belkin2006manifold}, blend the first and second options by combining domain knowledge, through an engineered distance metric/kernel, and unlabeled samples. Meanwhile, the representation learning community has pursued the second option, producing a number of methods like Deep Boltzmann Machines \cite{salakhutdinov2009deep}, Stacked Denoising Autoencoders \cite{vincent2010stacked}, and Sparse Coding \cite{raina2007self} that have proven effective for transfer from unlabeled data to tasks with few labeled samples.

While methods based on options (1) and (2) work well for a variety of tasks, they typically ignore the existence of large amounts of labeled data from different, but possibly related tasks, that may provide significant information about the task-of-interest (called \emph{target task}). The transfer learning community has explored this idea through a variety of methods, often termed ``supervised transfer''. The most common tools for supervised transfer come from multitask learning \cite{caruana1997multitask}, in which a fixed set of tasks is given a priori and the learner seeks a model that can \textit{generalize well to new \underline{samples}} from the given tasks. While multitask learning has been effective in many situations, it falls short in environments where new tasks are constantly arriving and one seeks to \emph{generalize well to new \underline{tasks}}, as is the case for our MLSP. This latter type of transfer is generally referred to as inductive bias learning \cite{baxter2000model}, lifelong learning \cite{thrun1995lifelong}, learning to learn \cite{thrun1996learning}, or never ending learning \cite{carlson2010toward}, among other names. Our work focuses on this setting.

In this paper, we present a method to learn in an environment of streaming tasks, like that faced by our MLSP. Our method operates on two components: a parametric representation (shared between tasks) and a collection of parametric function approximators (one per task).  We aim to learn the parameters of the representation such that its output would be a more effective input to new function approximators. Our method, called \textbf{LeaDR} (Learning Discriminative Representations), builds on the intuition that a good representation is one that allows transfer to new tasks with few labelled samples. We formalize this intuition through a novel objective function that minimizes an empirical proxy of the intra-task small-sample generalization error. This particular objective proves useful in forcing the representation to focus on small sample transfer to new tasks.


In developing our method, we tackle several challenges.  First, since tasks faced by our MLSP are not fixed, but rather dynamically defined implicitly through their labellings, new tasks cannot be mapped to previously seen tasks. Moreover, label information will be inconsistent among clients (i.e. one client's person of interest $\neq$ another client's person of interest). This aspect precludes transfer of label information among tasks. We will refer to this problem as \textit{task correspondence}. Another issue is scalability. Since we assume our learner operates in an environment where the number of streaming tasks is very large, memory and computational requirements should be sub-linear in the number of previously-presented tasks. Finally, as many problems of interest are inherently high dimensional, our learner must be efficient when dealing with high dimensional inputs. While some work has sought to address problems similar to our described MLSP~\cite{saha2011online, ruvolo2013ella}, these algorithms do not scale well with respect to the number of tasks and/or input dimensionality, and thus experimental results have been limited to low-dimensional multitask problems.

While our primary goal is to tackle the MLSP problem outlined above, our approach proves effective in a wider range of tasks. In our experimental results, we show that LeaDR can be used in three settings:  (1) to learn a good representation for achieving single task transfer (Sec. 5.1), where we outperform a state-of-the-art deep learner (Spike-and-Slab Sparse Coding) using only half the labelled samples on the target task, (2) to tackle standard multitask learning (Sec. 5.2), where we match or exceed performance of state-of-the-art multitask learning approaches despite not using task correspondence information, and (3) to solve the MLSP problem in high-dimensional input spaces (Sec. 5.3), with better scalability than previous lifelong learning approaches.

\section{A Machine Learning Service Provider: Problem Definition}
\label{chap:The Online Transfer Learning Problem}
Let our MLSP operate in an environment $\EE = (\QQ, \XX)$, with an input domain $\XX$ and a task distribution $\QQ$. $\QQ$ is a distribution over tasks $\TT_i$, where each task $\TT_i = (\YY_i, \DD_i, \LL_i, \GG_i)$ comprises:
\vspace{-\topsep}
\begin{enumerate}
\itemsep 0em
\item An output domain $\YY_i$.
\item A distribution $\DD_i$ over $\XX \times \YY_i$.
\item A non-negative loss function $\LL_i: \YY_i \times \YY_i \rightarrow \mathbb{R}^{+}$.
\item A generalization functional $\GG_i(h) = \mathbb{E}_{(x,y) \sim \DD_i} \LL_i(h(x), y)$.
\end{enumerate}
\vspace{-\topsep}
For the environments we consider, $\XX = \mathbb{R}^d$. For classification tasks, $\YY_i$ is the discrete space of relevant labels and for regression tasks, $\YY_i = \mathbb{R}$. The generalization functional $\GG_i(h)$ measures the true loss of the learned hypothesis $h$. For classification tasks, $\GG_i(h)$ is the expected misclassification rate of the classifier $h$ w.r.t.~$\DD_i$. For regression tasks, $\GG_i(h)$ is the expected error of the regressor $h$, w.r.t.~$\DD_i$. We do not restrict our definitions to specific loss functions as our model is able to accommodate different loss functions (e.g., logistic, ranking, RMSE, etc). 

Loosely speaking, our MLSP is a persistent machine learning agent faced with an environment $\EE$ and tasked with producing hypotheses $\hat{h}_i: \XX \rightarrow \YY_i$ for any tasks $\TT_i$ it encounters, so as to minimize $\mathbb{E}_{\TT_i \sim \QQ} \GG_i(\hat{h}_i)$. To produce a hypothesis $\hat{h}_i$ for task $\TT_i$, the agent first receives an $m$-sample $(X_i, Y_i) = \{(x_j, y_j)\}_{j=1}^m$ of training observations drawn from $\DD_i$. The agent then applies an algorithm $\AAA_i$ to the $m$-sample to produce a hypothesis $\hat{h}_i$ that minimizes a structured risk. We denote this process as $\hat{h}_i = \AAA_i(X_i, Y_i)$, with:
\begin{align*}
&\AAA_i(X_i, Y_i) = \argmin_{h \in \HH_i^{\AAA}} \frac{1}{m} \sum_{(x_j,y_j) \in (X_i, Y_i)} \LL_i(h(x_j), y_j) + \RR_i(h),
\end{align*}
where $\HH_i^{\AAA}$ gives the hypothesis space searched by $\AAA_i$, $\LL_i$ gives the loss function minimized by $\AAA_i$, and $\RR_i$ gives any regularization terms used by $\AAA_i$ to bias the search over $\HH_i^{\AAA}$. We let $\AAA_i$ minimize a surrogate loss $\LL_i^{\AAA}$, as the natural task loss $\LL_i$ may provide an intractable optimization objective.

Using the definitions presented thus far, our MLSP learning objective can be written as:
\begin{equation}
\minimize_{\theta \in \Theta} \mathbb{E}_{\TT_i \sim \QQ} \mathbb{E}_{(X_i, Y_i) \sim \DD_i} \GG_i(\AAA_i(X_i, Y_i)).\footnote{While we use $m$ for the number of samples used as input to each $\AAA_i$, in practice $m$ may differ across tasks.} 
\label{ll_obj1}
\end{equation}
The objective in~\eqref{ll_obj1} measures the ability of the per-task algorithms $\AAA_i$ to find hypotheses $\hat{h}_i = \AAA_i(X_i, Y_i)$ that generalize well w.r.t.~$\GG_i$, in a small-sample setting.  As outlined in the next section, in our framework, $\AAA_i$ is in fact structurally-biased by a common representation learned over all tasks.

\section{A Method for Learning Discriminative Representations (LeaDR)}
\label{chap:Method}

We now present a method for addressing the objective in \eqref{ll_obj1}.   The basic idea is to combine a single parametric feature extractor $f$ (shared between tasks), with an unbounded collection of trainable function approximators $\{..,,h_i,...\}$ (one per task).  Our approach aims to make the output of $f$ more effective as input to the algorithm $\AAA_i$ when training $h_i$ for task $\TT_i$. While the collection of function approximators may be unbounded, we do select each algorithm/approximator pair $\AAA_i/h_i$ from a finite set of methods, e.g.,~linear logistic regression for classification tasks, linear least-squares regression for regression tasks\footnote{While we focus in this paper on parametric function approximators, possible extensions to non-parametric function approximators as in \cite{snoek2011nonparametric} are also possible.}.

Using the notation defined in Section 2, our objective is given by:
\begin{equation}
\minimize_{\theta \in \Theta} \mathbb{E}_{\TT_i \sim \QQ} \mathbb{E}_{(X_i, Y_i) \sim \DD_i} \GG_i \left( \AAA_i(f_{\theta}(X_i), Y_i) \right),
\label{ldr_obj_true}
\end{equation}
where $f_{\theta}$ represents setting the parameters of the parametric function approximator $f$ to $\theta$. This objective is an instance of the objective in \eqref{ll_obj1} in which structure sharing among the per-task algorithms---i.e.,~the $\AAA_i$ in \eqref{ll_obj1}---is accomplished by using the same parameterized feature extractor $f_{\theta}$ to preprocess inputs to each $\AAA_i$.

The particular novelty in our method is that we explicitly minimize an empirical proxy for the expected per-task small-sample generalization errors given by $\GG_i \left( \AAA_i(f(X_i), Y_i) \right)$. In contrast, typical approaches to multitask learning simultaneously learn a feature extractor $f_{\theta}$ and a collection of task-adapted functions $\hat{h}_i$ for a fixed set of tasks $\{\TT_1,...,\TT_n\}$. This approach seeks an $f$ such that there exist functions $\hat{h}_i$ with small error, w.r.t.~$\LL_i$, on the training sets available for each $\TT_i$.



\begin{figure}
  \centering
  \includegraphics[scale=.7]{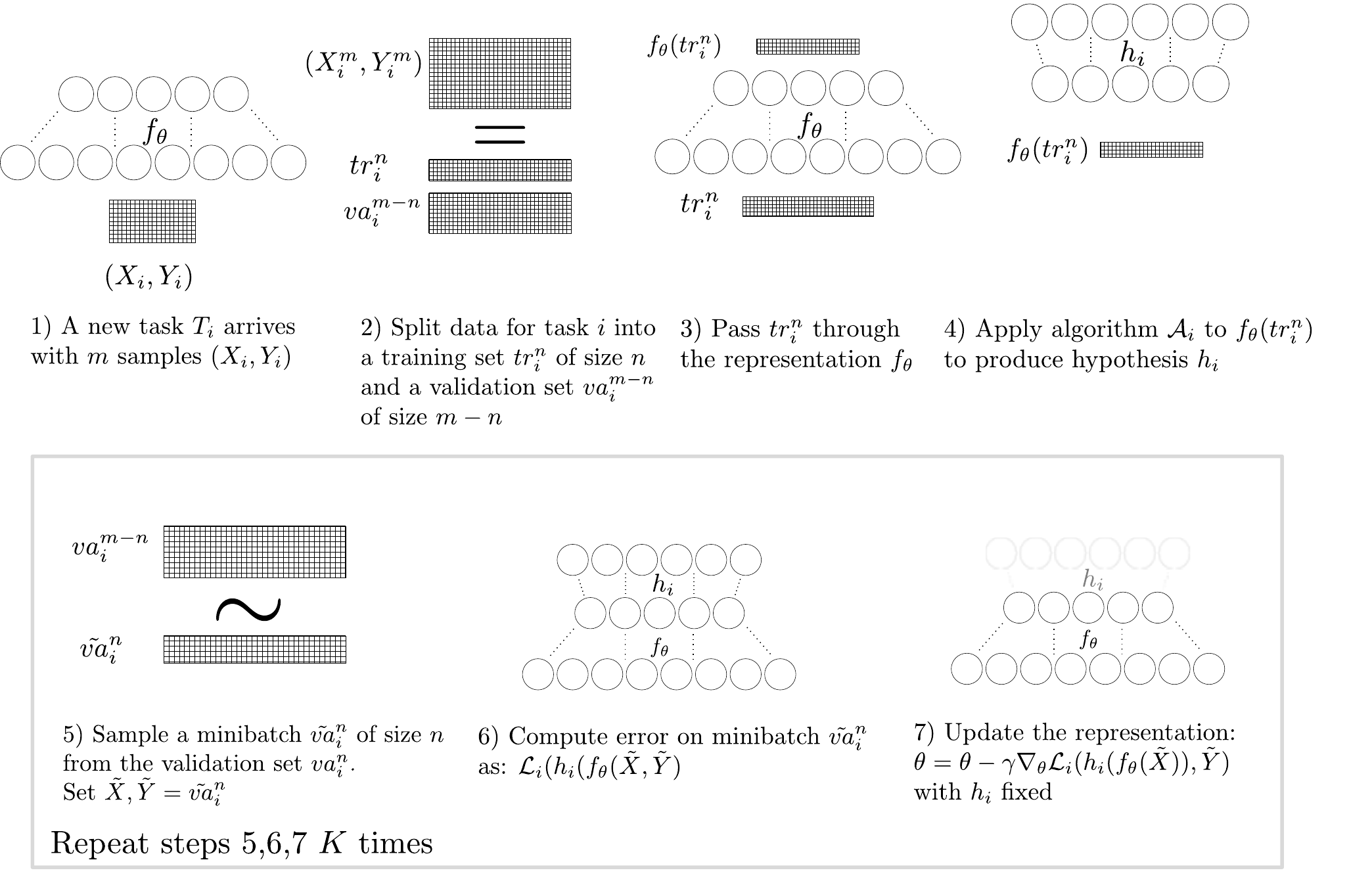}
  \caption{Key Steps for Learning Discriminative Representations (LeaDR) for Transfer}
\end{figure}

Given $T$ sets of $m$-samples $\{(X_1, Y_1),...,(X_T,Y_T)\}$ drawn from the environment, our method for Learning Discriminative Representations (called LeaDR) optimizes the following empirical approximation of \eqref{ldr_obj_true}:
\begin{equation}
\minimize_{\theta \in \Theta} \frac{1}{T} \sum_{i=1}^{T} \hat{\GG}_i(\AAA_i, (f_{\theta}(X_i), Y_i)),
\end{equation}
where $\hat{\GG}_i$ is an empirical estimate of generalization functional on task $i$. To define $\hat{\GG}_i$, we first define a process for sampling a pseudo-training/validation set pair $(tr_i^n, va_i^{m-n})$ from the $m$-sample $(X_i, Y_i)$ available for $\TT_i$. We sample the pseudo-train/validate split $(tr_i^n, va_i^{m-n})$ by randomly sampling $n < m$ observations $(x_i,y_i)$ for $tr_i^n$, from $(X_i, Y_i)$, and then let $va_i^{m-n}$ contain the remaining $m - n$ observations in $(X_i, Y_i)$.
Now, we compute $\hat{\GG}_i$ as follows:
\begin{align*}
\hat{\GG}_i(\AAA_i, (f_{\theta}(X_i, Y_i))) =
\mathbb{E}_{(tr_i^n, va_i) \sim (X_i, Y_i)} \left[ \sum_{(x_j, y_j) \in va_i} \LL_i(\hat{h}_i^n(f_{\theta}(x_j)), y_j) \right],
\end{align*}
in which $\hat{h}_i^n = \AAA_i(f_\theta(tr_i^n))$ indicates the hypothesis produced by applying algorithm $\AAA_i$ to the pseudo-training set $tr_i^n$. 
While $tr_i^n$ is used to train the task-specific hypothesis,  $va_i^n$ is used to train the intra-task representation $f_\theta$.

\begin{algorithm}[tb]
   \caption{Learning Discriminative Representations (LeaDR) for Transfer}
   \label{alg:algorithm_1}
\begin{algorithmic}[1]
   \STATE {\bfseries Inputs:} Source environment $\EE = (\XX, \QQ)$, pseudo-training batch size $n$, updates per task sample $K$, representation learner $f_{\theta}$ with initial parameters $\theta=\theta_0$, learning rate $\gamma$.
   \WHILE{ $\TT_i \sim \QQ$ is requested } 
   \STATE Observe $m$ training examples, $(X_i,Y_i) \sim \DD_i$
   \STATE Sample train/validate split $(tr_i^n, va_i^{m-n}) \sim (X_i,Y_i)$
   \STATE Create an appropriate algorithm/approximator pair $\AAA_i/h_i$
   \STATE Apply $\AAA_i$ to $tr_i^n$ to get $\hat{h}_i^n$.
   \FOR{$k=1$ {\bfseries to} $K$}   
	   \STATE Sample a minibatch $\tilde{va}_i^n$ from $va_i^n$.
	   \STATE Define $(\tilde{X}, \tilde{Y})$ such that $\tilde{va}_i^n = (\tilde{X}, \tilde{Y})$.
	   \STATE Let $\theta := \theta - \gamma \nabla_{\theta} \LL_i(\hat{h}_i^n(f_{\theta}(\tilde{X})), \tilde{Y})$
   \ENDFOR
   \ENDWHILE
   \STATE {\bfseries Output:} Learned representation $f_\theta$.
   
\end{algorithmic}
\end{algorithm}

Algorithm \ref{alg:algorithm_1} describes concretely the steps of our algorithm. Figure 1 presents a visual depiction of each step to further elucidate the algorithm. Note that in the loop on line 6, after altering the representation once, the optimal function approximator given the representation changes. For computational reasons, we approximate the new optimal function with the one acquired before updating the representation.



Our method has the following key properties:
\vspace{-\topsep}
\begin{enumerate}
\itemsep 0em
\item \textbf{Scalability}: LeaDR is scalable with respect to the number of tasks as computation per-task and memory requirements for all tasks are both $O(1)$. It is also scalable with respect to the dimensionality of the input space and the extracted feature space.



\item \textbf{Flexibility}:  LeaDR is modular, and is essentially a meta-algorithm that can easily accommodating different feature extractors, $f_\theta$ (e.g., boosted stumps, convolutional nets), as well as different function approximators, $\hat{h}_i$ (e.g., regressors, classifiers).
\item \textbf{Streaming}:  Tasks can be presented to our method sequentially, and need not be remembered explicitly (only through the parameters of the shared representation).
\end{enumerate} 
\vspace{-\topsep}

\section{Related Works}
\label{chap:Related Works}
Our work relates to several sub-fields.  Since our approach focuses primarily on transfer between tasks using labelled data, we focus here on literature related to supervised transfer. For a more extensive survey of transfer learning, see~\cite{pan2010survey}.   Ideas grouped under supervised transfer can be categorized into three sub-fields: \emph{multitask learning}, \emph{transductive learning} and \emph{lifelong learning}.

In \emph{multitask learning}, the goal is to solve a number of fixed problems simultaneously, with the hope that by sharing information between tasks, we can achieve better solutions for the problems considered~\cite{caruana1997multitask}. Many structures have been investigated to share information across tasks, including neural networks \cite{baxter1995learning,caruana1997multitask}, distance metrics \cite{thrun1996learning}, Bayesian priors \cite{xue2007multi}, sparse code dictionaries \cite{kumar2012learning}, and others. Overall, multitask learning assumes the marginal distributions of samples seen during testing to be the same as those during training, and that with each sample, a specific task id is given. The Online MultiTask Learning (OMTL) framework~\cite{saha2011online} can be used to tackle a stream of related tasks\footnote{The word ``online'' in OMTL refers to samples from fixed tasks arriving in an online manner. This is distinct from the stream of \emph{tasks} faced by our MLSP.}. However OMTL does not scale to environments with large number of tasks as it requires maintaining a $T \times T$ task-relatedness matrix ($T=$number of tasks). 
In \emph{transductive learning} \cite{daume2006domain,zadrozny2004learning}, the source and target tasks are also the same, however the marginal distributions differ between the training and testing samples. The focus of transductive learning has been on correcting for this difference, to improve performance in target domains.

\emph{Lifelong learning} differs from the previous two sub-fields in that target problems are assumed to be \text{distinct} from training problems~\cite{baxter2000model,thrun1995lifelong,baxter1995learning}. In general, methods used for lifelong learning can be applied in situations where multitask or transductive learning are used, but not the reverse, since lifelong learning relaxes the task correspondence assumption.   Theoretical foundations for lifelong learning were established by Baxter~\cite{baxter2000model}, wherein PAC bounds for a setting similar to the one in Sec.~\ref{chap:The Online Transfer Learning Problem} were presented. The algorithm in \cite{baxter2000model} uses a two-part model to optimize a multitask learning objective. This model however, consumes an amount of memory linear in the number of tasks. Baxter's method extends to learning deep representations \cite{collobert2008unified}.


The best method we are aware of to tackle lifelong learning over large number of tasks is ELLA~\cite{ruvolo2013ella}, which builds on a class of methods in multitask learning where multiple tasks share a common basis. The main idea is to represent the parameters of the predictive function $f_w$ for a new task using a sparse code over dictionary elements, i.e.~$f_w(x) = (L s)^{\top} x$, where $||s||_1$ is small and $w = L s$. While ELLA is a lifelong learner, consuming $O(T)$ time and $O(1)$ space for $T$ tasks, it is inefficient in domains with high dimensional input spaces, requiring $O(d^3k^2)$ computation time, where $d$ is the dimensionality of the data and $k$ is the number of bases in the sparse code. 

While our MLSP is primarily focused on learning representations for inter-task transfer from a large stream of tasks, the LeaDR approach we propose can also improve transfer to new tasks when learning from just a single encountered task. As such, our work is connected to recent literature in computer vision where deep neural networks are trained on a single highly multiclass task with large amounts of labeled data, after which the features learned by the network are applied to tasks other than the training task, for which few labeled samples are available \cite{razavian2014cnn,Jia13caffe,donahue2013decaf}. Using these deeply and supervisedly learned features has led to a rapid advance in the state-of-the-art for a number of domains \cite{razavian2014cnn}. Our method can be seen as a new approach to training such networks, aimed at improving transfer to new tasks with very small training sets.

\section{Empirical Evaluation}
\label{chap:Evaluation}
\vspace{-3mm}
We evaluate our method in three different contexts, selected to reflect a range of transfer settings. 
\vspace{-3mm}
\subsection{Representation Learning for Single Task Transfer}
First we consider the single task transfer case, with the goal of showing that LeaDR provides a robust solution for supervised transfer to new tasks. This experiment targets the NIPS 2011 transfer learning challenge \cite{nips_challenge_2011} which was proposed during the NIPS workshop on Learning Hierarchical Representations. Data is available as follows: a first dataset of 50,000 $32 \times 32 \times 3$ images labelled into 100 classes is given (namely, the CIFAR-100 dataset \cite{krizhevsky2009learning}), along with a second dataset of 100,000 unlabelled $32 \times 32 \times 3$ images. The target domain is a 10-way classification problem, for which only 120 labelled training examples are given; each of the classes in this target domain is distinct from those seen in the labelled dataset. The goal is to create a method that performs well on the target domain's test set of another 2542 samples. The best previously published result for this challenge was Spike and Slab Sparse Coding (S3C) \cite{goodfellow2012large} which ignores the labels on CIFAR-100 and uses it just as unlabelled data. After training, the representation learned by S3C is used to extract features, on which a linear classifier is trained using the 120 samples for the target task.

To instantiate the LeaDR framework, we used a convolutional neural network for the representation, and logistic regression (trained by gradient descent) for the function approximator.  A stream of training tasks was simulated by sampling different 20-way classification problems from the CIFAR-100 dataset. After training the representation in this way, we used the learned $f_{theta}$ to extract features on the target task, after which, we trained $\hat{h}_i$ for the target task using the 120 samples (or less). The test set for the target task (2542 samples) was projected through the shared learned representation and then through the task-specific hypothesis to evaluate performance. We compared our method to simply learning a representation using a Convolutional Neural Network trained by Backpropagation on the full CIFAR-100 dataset, where the penultimate layer of the ConvNet is replaced by a task-specific logistic regression as in our method. Note that for vision tasks, such a convolutional architecture presents a remarkably strong baseline~\cite{razavian2014cnn}. Both methods used the same convolutional architecture for their representations, and ignored the unlabelled data. More implementation details are in the supplementary.

\begin{wrapfigure}{r}{0.52\textwidth}
  \includegraphics[scale=0.8]{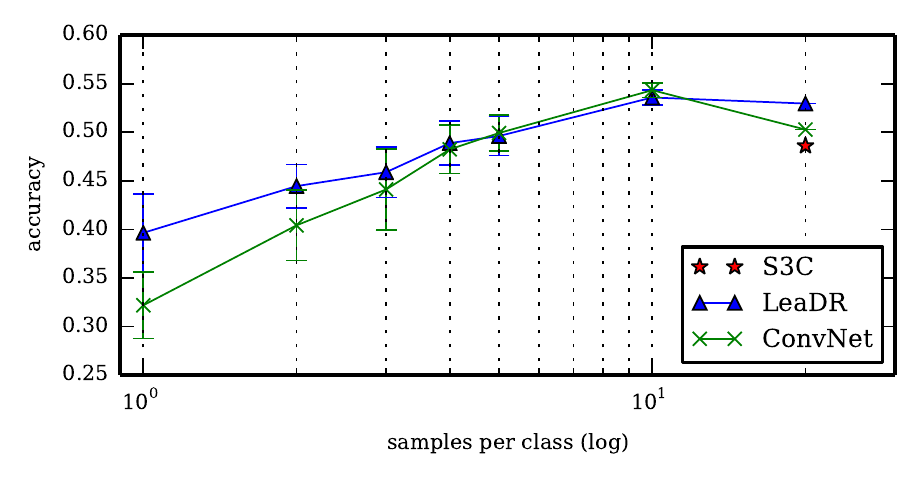}
  \caption{NIPS 2011 transfer learning challenge}
\end{wrapfigure}

Figure 2 shows the performance of LeaDR compared to the standard ConvNet with Backprop, as the number of samples per class increases. We see that LeaDR has an advantage when new tasks have less than 3 samples per class (30 in total). In the ``one shot'' case, LeaDR presents a gain of about 8\% compared to the ConvNet.  At 5 labelled samples per class (50 in total), we see that representations trained in a supervised way start to outperform unsupervised pre-training (S3C, which uses all 120 labels for the target task). Surprisingly, for the ConvNet with Backprop, performance decreases when we train on the whole 120 samples. This may be due to the uneven distribution of 120 samples among the 10 classes on the test problem. The performance of our algorithm in small sample cases showcases the advantage of minimizing empirical generalization, as opposed to standard loss.

\subsection{Multitask Learning Experiments}
Second we focus on the standard multitask learning case, where the training tasks are fixed and testing tasks are constrained to be from the same set of tasks.   We show that even though LeaDR was not designed for standard multitask learning, and does not exploit task correspondence, it compares favorably to state-of-the-art multitask learners on two problems from this sub-field:  a classification domain called \emph{Landmine} \cite{xue2007multi} and a regression domain called \emph{London schools} \cite{kumar2012learning}; both datasets have a few dozen tasks, and a relatively low-dimensional input space.


We compare our method to three algorithms designed for multitask and lifelong learning: GOMTL \cite{kumar2012learning}, OMTL \cite{saha2011online} and ELLA \cite{ruvolo2013ella}. We also include comparison to standard single task learning (STL) as a baseline: logistic regression for the Landmine (classification) problem and random forests for the London schools (regression) problem. We apply the experimental methodology used in~\cite{ruvolo2013ella}, where data for each task (within a domain) is split 50/50 between training/testing sets.  We measure error on Landmine (classification) in terms of Area Under the Curve (AUC)\footnote{The choice of AUC as opposed to standard misclassification error is to conform with \cite{ruvolo2013ella}. In our experiments, misclassification error was about the same for all the methods, including STL. In our opinion, this suggests the Landmine dataset may not be a great domain for multitask learning.}. For the London schools (regression), we measure error in RMSE.   More details on the datasets, training process and experimental setup are in the supplemental material.

Table 1 contains the results, showing that LeaDR compares favourably to other state-of-the-art algorithms on these standard multitask datasets. We note that due to the simplicity of these domains, in terms of task specification and input dimension, most methods perform rather similarly, and there is little room to gain relative to single task learning. Our final experimental setting is intended to move past some of these limitations, and examine performance in a setting that more closely approaches the MLSP problem, including streaming (unknown) tasks and high dimensional input spaces.
\vspace{-3mm}
\begin{table}[!htp]
\caption{Performance of LeaDR against state-of-the-art multitask learning algorithms. Let $d$ be the dimensionality of the input space, $|T|$ the number of tasks in the dataset, $N$ the number of available examples in the dataset.}
\vspace{2mm}
\centering
    \begin{tabular}{l|c|c|c|c|c|c}	
    \textbf{Dataset}        & \textbf{Error Type} & \textbf{OMTL} & \textbf{GOMTL} & \centering \textbf{ELLA}  & \textbf{STL} & \textbf{LeaDR} \\ \hline
     \textbf{Landmine} &&&&&&\\
     $d=9$, $|T|=29$     & AUC & 0.63   & 0.78 & 0.776 & 0.76 &  0.78           \\
     $N=14,820$ &&&&&&\\
     \textbf{London Schools} &&&&&&\\
     $d=27$, $|T|=139$  & RMSE & N/A  & 10.10 & 10.20 & 11.06  & 10.08      \\
     $N=15,362$&&&&&&\\
     \hline
    \end{tabular}
\end{table}
\vspace{-3mm}
\subsection{Lifelong Learning Experiments}


In our final set of experiments, we tackle the full lifelong learning problem, where we investigate how LeaDR performs when presented with a large set of streaming tasks, and where testing is done on different problems than seen during training, as in the MLSP setting.  We consider variations on two standard ML domains:  the 20 Newsgroups text classification domain~\cite{20_newsgroups}, and the above-mentioned CIFAR-100 image classification domain~\cite{krizhevsky2009learning}.

We simulate the online streaming of tasks by sampling as follows.  We sample 1000 random (training) 5-way classification problems from the respective set of classes, and apply both stages of LeaDR (representation + fn approx.) to these tasks.  We then fix the representation learner, $f_\theta$and present 100 new (testing) 5-way classification tasks. We optimize a task-specific hypothesis, $\hat{h}_i$, for each of these tasks using a very small set of labelled samples (processed through the learned representation stage). Finally we test the accuracy of each task-specific hypothesis on a test set for that particular task.  Samples used for training the representation, training the hypotheses for the test tasks, and for measuring the accuracy on each test task, come from three disjoint sets.

We instantiate LeaDR with a two layer neural net for the representation learning, and logistic regression for the task-specific function approximators. We compare our method to ELLA \cite{ruvolo2013ella} and a single-task multinomial logistic regression (where training data from all tasks is mixed in a single batch).  Other algorithms \cite{saha2011online,kumar2012learning} were not included because they cannot scale to this many training tasks.  Since ELLA cannot efficiently handle problems with high dimensional input spaces, we consider low and medium dimensional projections of the input spaces, in addition to the (natural) high dimensional input space (we cannot include ELLA in this last case).  For 20 Newsgroups, the high dimensional input space assumes each document is represented by a vector of counts for the 2000 most-common non-stop words; the low (d=10) and medium (d=100) input spaces are created using LDA \cite{blei2003latent} over these word vectors.  For CIFAR-100, the high dim. input space uses raw pixel values, whereas the low and medium input spaces are obtained by projecting this down using PCA.  We also include results for LeaDR on CIFAR-100 with high dimensional inputs where the representation learner is replaced by a ConvNet (rather than a two layer NN).  

Figure 3 presents the accuracy of each method as the number of samples for testing tasks increases.   We observe that LeaDR consistently outperforms ELLA, regardless of available sample size, and also has much lower variance.  Surprisingly, single-task logistic regression sometimes outperforms ELLA; possibly because ELLA cannot learn effective transfer representations of these domains with its particular sparse coding architecture. LeaDR is particularly strong when tackling high dimensional input spaces with very few samples. To the best of our knowledge, this paper is the first to present results in this particular setting.

%

\vspace{-3mm}
\begin{figure}[hctp]
\hspace*{-.2in}
 \subfigure[c][Results on CIFAR-100 ]
 {\includegraphics[width=15cm]{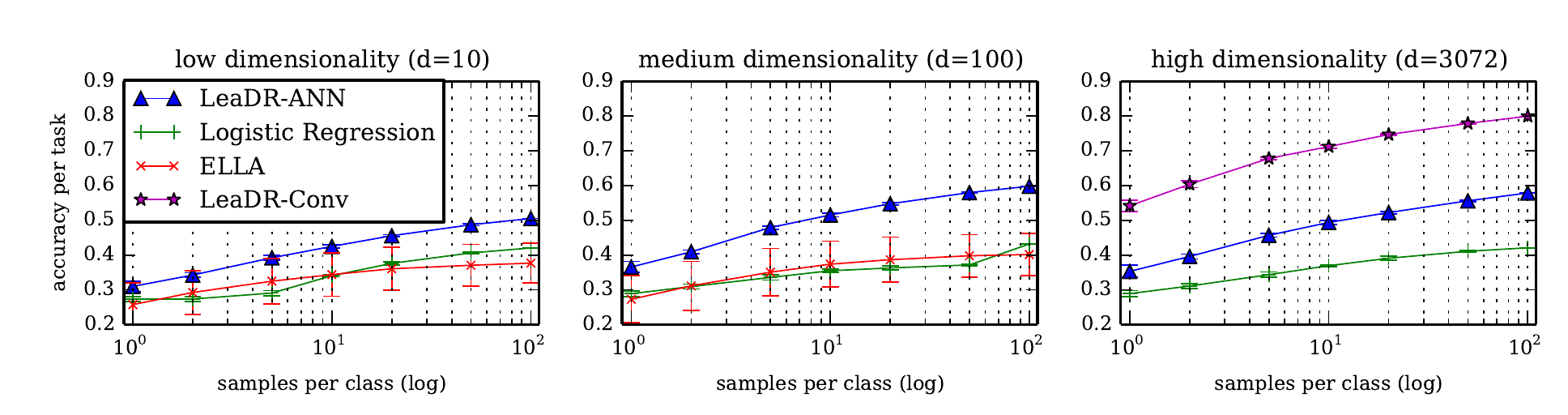}}

\hspace*{-.1in}
 \subfigure[c][Results on 20-newsgroups]{\includegraphics[width=15cm]{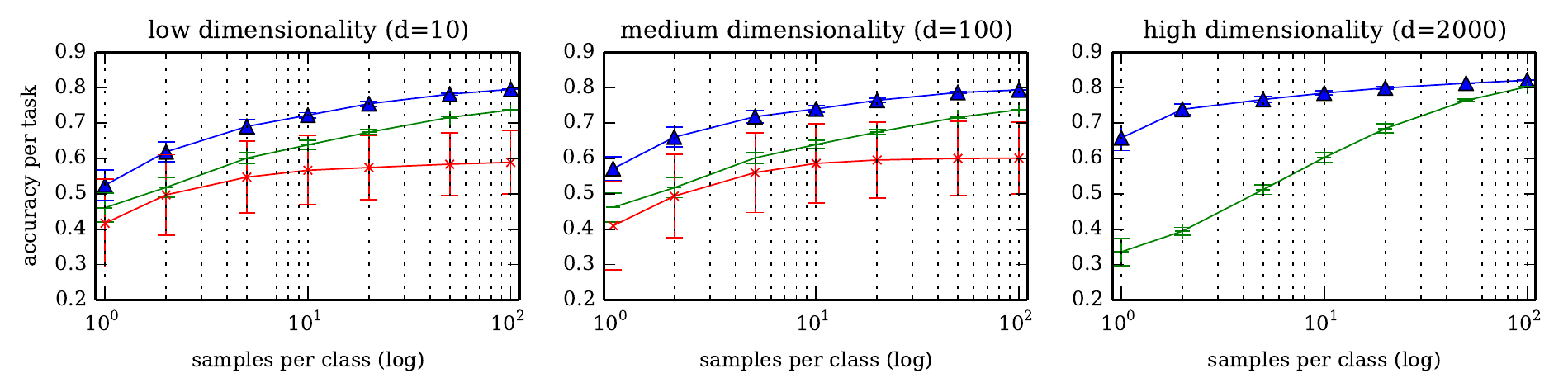}}
 \vspace{-3mm}
 \caption{Results for the lifelong learning experiments. $d$ is the number of dimensions in the input.} Each point is averaged over the 100 different test problems, and over 10 different sample sets for each test task.
\end{figure}


\vspace{-5mm}
\section{Discussion}
\vspace{-3mm}
\label{chap:Discussion}

In this paper, we formalized an interesting and challenging learning problem, termed the Machine Learning Service Provider, that deals with the problem of rapidly creating accurate learners in environments with streaming (related but non-identical) tasks.  We then presented an algorithmic framework (LeaDR) that tackles this problem in a flexible and scalable way. A particular novelty of our method, is to learn the feature representation by minimizing an empirical proxy of the inter-task generalization error. This particular objective proves useful in many scenarios when the goal is to use the learned representation  for transfer. Our method presents several desirable properties, including flexibility, scalability. The flexibility of our approach is in its modularity, as both the representation and task-specific function approximators can be changed to suit the input domain and tasks at hand. In terms of scalability, our framework allows scaling multitask and lifelong learning to high dimensional, streaming tasks regimes. Empirically, we verified these claims in three relevant contexts: single-task transfer, where our method was shown to be better at learning convolutional supervised representations than standard backpropagation; Multitask learning: where our algorithm matches our outperforms state-of-the-art multitask learners and finally, lifelong learning, where our algorithm outperforms other methods when facing a large stream of tasks with high dimensional inputs.

\pagebreak
\small
{
\bibliographystyle{splncs}
\bibliography{ldr}
}

\end{document}